\documentclass[10pt,twocolumn,letterpaper]{article}

\usepackage[pagenumbers]{cvpr} 

\usepackage{graphicx}
\usepackage{amsmath}
\usepackage{amssymb}
\usepackage{booktabs}

\usepackage{makecell}
\usepackage{multirow}

\usepackage[pagebackref,breaklinks,colorlinks]{hyperref}

\usepackage[capitalize]{cleveref}
\crefname{section}{Sec.}{Secs.}
\Crefname{section}{Section}{Sections}
\Crefname{table}{Table}{Tables}
\crefname{table}{Tab.}{Tabs.}

\def\cvprPaperID{***} 
\def\confName{CVPR}
\def\confYear{2023}

\newcommand{\ignorethis}[1]{}

\begin{document}

\title{PointOcc: Cylindrical Tri-Perspective View for Point-based 3D Semantic Occupancy Prediction}

\newcommand*\samethanks[1][\value{footnote}]{\footnotemark[#1]}
\author{Sicheng Zuo\footnotetext{sadfsadfa}\thanks{Equal contribution.}\quad Wenzhao Zheng\samethanks\quad Yuanhui Huang\quad Jie Zhou\quad Jiwen Lu\thanks{Corresponding author.} \\
Beijing National Research Center for Information Science and Technology, China \\
Department of Automation, Tsinghua University, China \\
\texttt{\{zuosc19,zhengwz18,huangyh22\}@mails.tsinghua.edu.cn;} \\
\texttt{\{jzhou,lujiwen\}@tsinghua.edu.cn}
}

\maketitle

\newcommand\blfootnote[1]{%
  \begingroup
  \renewcommand\thefootnote{}\footnote{#1}%
  \addtocounter{footnote}{-1}%
  \endgroup
}

\begin{abstract}
Semantic segmentation in autonomous driving has been undergoing an evolution from sparse point segmentation to dense voxel segmentation, where the objective is to predict the semantic occupancy of each voxel in the concerned 3D space.
The dense nature of the prediction space has rendered existing efficient 2D-projection-based methods (e.g., bird's eye view, range view, etc.) ineffective, as they can only describe a subspace of the 3D scene.
To address this, we propose a cylindrical tri-perspective view to represent point clouds effectively and comprehensively and a PointOcc model to process them efficiently.
Considering the distance distribution of LiDAR point clouds, we construct the tri-perspective view in the cylindrical coordinate system for more fine-grained modeling of nearer areas.
We employ spatial group pooling to maintain structural details during projection and adopt 2D backbones to efficiently process each TPV plane. 
Finally, we obtain the features of each point by aggregating its projected features on each of the processed TPV planes without the need for any post-processing.
Extensive experiments on both 3D occupancy prediction and LiDAR segmentation benchmarks demonstrate that the proposed PointOcc achieves state-of-the-art performance with much faster speed.
Specifically, despite only using LiDAR, PointOcc significantly outperforms all other methods, including multi-modal methods, with a large margin on the OpenOccupancy benchmark.
Code: \url{https://github.com/wzzheng/PointOcc}.

\end{abstract}
\section{Introduction}
Accurately and comprehensively perceiving the 3D environment is a crucial aspect of the autonomous driving system.
With the ability to actively detect the 3D structural information of a scene, LiDAR has become the mainstream sensor for most autonomous driving vehicles, where LiDAR-based models have dominated most of the main perception tasks for autonomous driving, including 3D object detection~\cite{pointpillars, beverse, pv-rcnn}, semantic segmentation~\cite{gfnet, pvkd, sphericaltr}, and object tracking~\cite{transfusion, voxelnext, bevfusion, bytetrack}.

\begin{figure}[t] 
\centering
\includegraphics[width=0.475\textwidth]{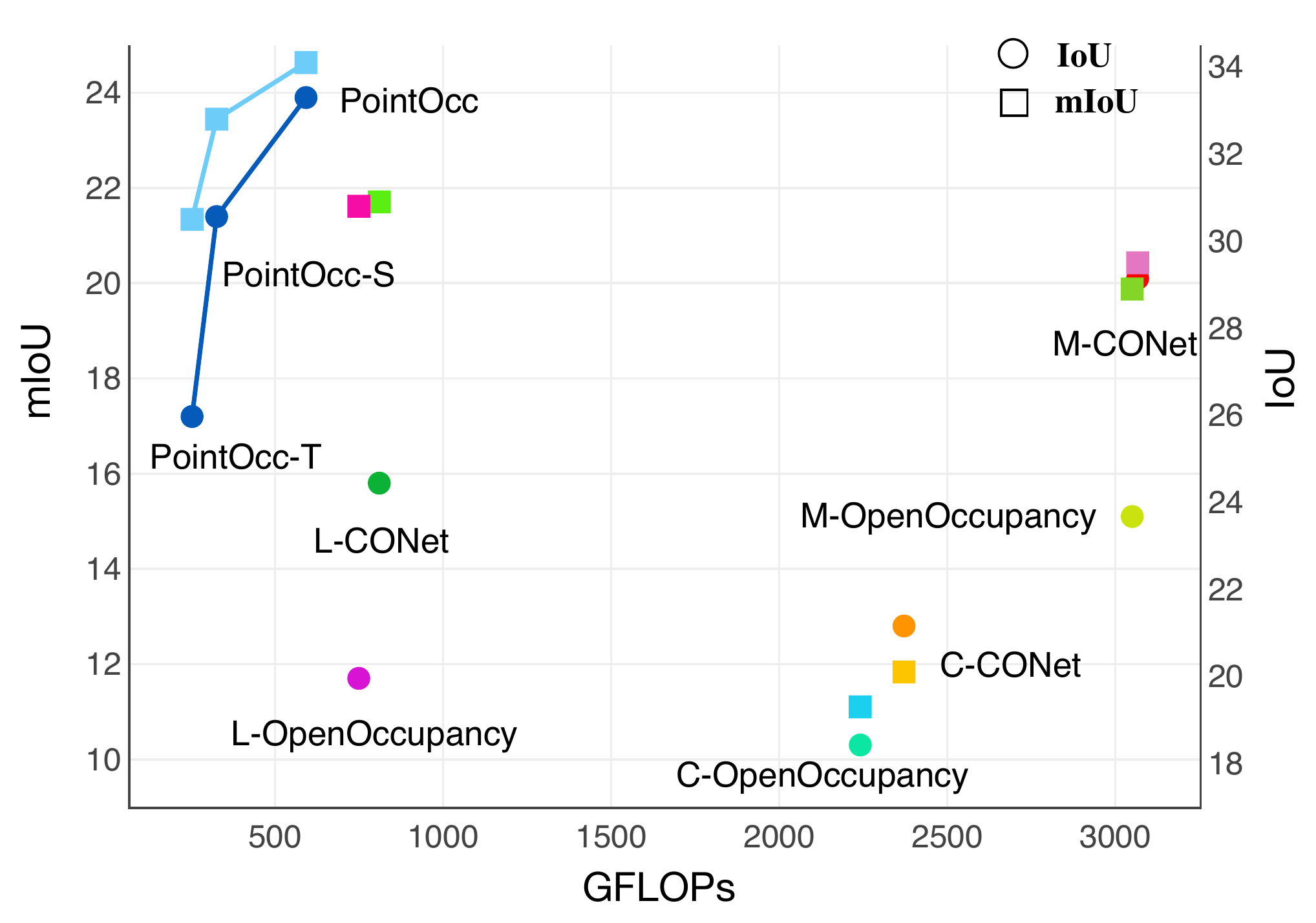}
	\vspace{-7mm}
\caption{Performance comparisons of the proposed PointOcc. 
Our PointOcc outperforms the best-performing multi-modal method by a large margin with much faster speed on the 3D semantic occupancy prediction task.
}
\label{fig:overview}
	\vspace{-6mm}
\end{figure}

LiDAR semantic segmentation plays an important role in autonomous driving perception with the objective to predict the category of each point.
One straightforward way is to use discretized voxels to represent point clouds and perform 3D operations on the voxels~\cite{voxelnext,voxelnet,spvnas,voxeltr}.
Considering the sparse nature of the LiDAR point clouds, some other LiDAR segmentation methods first perform 2D projection on the point clouds to obtain range images or bird's eye view (BEV) images as inputs to 2D backbones~\cite{polarnet, rangenet++, salsanext, rangevit, amvnet}.
Though fast in speed, they usually underperform the voxel-based methods due to the information loss of the 3D-to-2D projection and require expensive post-processing to restore the 3D structure~\cite{salsanext, rangenet++, rangevit}.
On the other hand, the semantic segmentation of sparse point clouds can only provide partial semantic descriptions of the 3D surroundings and cannot generalize well to arbitrary-shaped obstacles, which is the key to the safety of autonomous driving vehicles.
Therefore, the autonomous driving community has recently shifted its focus to the 3D semantic occupancy prediction problem~\cite{occ3d, openoccupancy, monoscene, tpvformer, wei2023surroundocc}, which aims to simultaneously predict the occupancy and semantic label of each voxel in the surrounding 3D space.
It is not trivial to adapt existing 2D-projection-based methods to 3D semantic occupancy prediction due to its dense output space.

In this paper, we propose PointOcc, an efficient 2D-projection-based model that can be applied to 3D semantic occupancy prediction without any requirement for post-processing and achieves even better results than 3D models, as shown in Figure~\ref{fig:overview}.
Motivated by the TPVFormer~\cite{tpvformer} for vision-based 3D semantic occupancy prediction, we represent the 3D scene using three 2D planes, each of which provides descriptions from one perspective.
While TPVFormer employs three perpendicular planes in the Cartesian coordinate system, we empirically find that they cannot well describe the LiDAR point clouds due to the non-uniform distribution of point clouds (i.e., the sparsity of the points depends on their distances to the ego car).
Therefore, we propose a cylindrical tri-perspective view (Cylindrical TPV) as a novel and effective representation of point clouds, which constructs TPV in the cylindrical coordinate system.
Cylindrical TPV naturally assigns nearer areas with a larger resolution and thus is suitable for LiDAR point clouds.
To transform point clouds into Cylindrical TPV, we employ spatial group pooling to project point clouds to each of the TPV planes with better preservation of 3D information. 
Having obtained the TPV planes, we can adopt any 2D backbones to perceive them without expensive 3D operations.
Finally, we model each point (or voxel) in the 3D space by summing its projected features on each of the processed planes. 
We then use a simple classification head to classify each 3D point feature without any need for post-processing.
Though each Cylindrical TPV plane can only provide a 2D description, they represent a 3D scene from complementary perspectives and collaborate to comprehensively model 3D information.

We conducted extensive experiments on the OpenOccupancy~\cite{openoccupancy} benchmark for 3D semantic occupancy prediction and the Panoptic nuScenes~\cite{nuscenes} benchmark for LiDAR segmentation.
We find that despite sharing the fast speed of 2D-projection-based methods, PointOcc can readily benefit from 2D-pretrained image backbones to achieve even better performance than the less efficient voxel-based methods.
Our PointOcc outperforms all other 2D-projection methods~\cite{salsanext, rangenet++, rangevit} on the LiDAR segmentation task and is comparable to voxel-based methods~\cite{cylinder3D, rpvnet, lidarmultinet} without any post-processing.
As the first 2D-projection-based method on the 3D semantic occupancy prediction task, PointOcc significantly outperforms all other methods by a large margin with a much faster speed.
Despite only using LiDAR as inputs, PointOcc is even better than the best multi-modal (LiDAR \& Camera) method~\cite{openoccupancy} by 4.6 IoU and 3.8 mIoU, demonstrating its advantage and potential.

\section{Related Work}

\textbf{LiDAR Segmentation.}
LiDAR segmentation aims to assign a semantic label to each point in the LiDAR point clouds, which serves as the basic task for LiDAR semantic perception.
Existing state-of-the-art methods~\cite{spvnas, af2s3net, lidarmultinet, sphericaltr} usually split the 3D space into uniform voxels and process voxels with 3D convolutional networks, resulting in dense cubic features to describe the 3D scenes~\cite{openoccupancy}.
Despite the strong performance of voxel-based methods, they usually suffer from significant computational and storage burdens due to the complex 3D operations on a vast number of voxels~\cite{voxelnet, voxelnext, voxeltr, sparseconv}.
Therefore, considering the sparse and varying density of LiDAR point clouds, other methods explored projecting point clouds onto a 2D plane and using 2D backbones to process pseudo-image features~\cite{polarnet, rangenet++, rangevit, amvnet}, and can thus greatly reduce computational and storage overhead. 
BEV-based methods compress 3D space along the z-axis and only encode point features on the ground plane~\cite{polarnet}. 
Range-view-based methods transform point clouds to spherical coordinates and project them along the radial axis to obtain the 2D range view features~\cite{rangenet++,salsanext, rangevit}. 
However, the loss of structural information during the 3D-2D projection results in lower performance compared to voxel-based methods. 
In addition, 2D projection-based methods require complicated post-processing techniques~\cite{kprnet, crf, kpconv} to restore 3D structural information, making it difficult to deploy them in real-world applications.

Recently, TPVFormer~\cite{tpvformer} proposed a Tri-Perspective View (TPV) representation for vision-based 3D perception, which uses three orthogonal complementary 2D planes to model the 3D scene. 
Due to the complementary properties of the three planes, TPV representation can effectively restore the 3D structure while maintaining efficiency. 
Still, TPVFormer only employs TPV to model already extracted image features.
It remains unknown how to transform LiDAR point clouds into TPV and how to process them using 2D image backbones.
To the best of our knowledge, we are the first to effectively apply TPV to LiDAR-based 3D perception.
We further propose a Cylindrical TPV representation to adapt to LiDAR point clouds and employ a spatial group pooling method to effectively transform LiDAR into TPV with minimum information loss.

\textbf{3D Occupancy Prediction.}
The LiDAR segmentation task predicts labels only for the sparse lidar points, and thus cannot provide a comprehensive and fine-grained description of the 3D scene, which is essential for autonomous driving systems. 
To address this, recent methods~\cite{s3cnet, scpnet, monoscene} started exploring dense semantic predictions for all the voxels in the surrounding space, formulated as the 3D occupancy prediction task~\cite{openoccupancy, occ3d}.
This task requires a simultaneous prediction of the occupancy status for all voxels and semantic labels for occupied voxels.
By modeling densely distributed voxels in 3D space, the 3D occupancy prediction task achieves a fine-grained and complete perception of the full space. 
Therefore, 3D occupancy prediction is a promising and challenging task in the field of autonomous driving perception.
The dense prediction space motivates most existing works to adopt a voxel-based model~\cite{udnet, s3cnet, js3c, scpnet, openoccupancy}.
For example, SCPNet~\cite{scpnet} utilizes a 3D completion network without downsampling and distills rich knowledge from the multi-frame model. 
L-CONet~\cite{openoccupancy} introduces a coarse-to-fine supervision strategy to reduce heavy computational burdens.

Though voxel-based methods can model fine-grained structures in 3D space, the dense voxel features introduce a huge storage overhead~\cite{scpnet, openoccupancy}. 
Due to the computation resource limitation, only low-resolution occupancy predictions can be obtained, which severely degrades the performance. 
However, it is non-trivial to directly adapt existing 2D-prediction-based lidar segmentation methods to 3D occupancy prediction due to the difficulty to recover 3D dense features from 2D projected features.
In this paper, we address this problem by transforming LiDAR point clouds to the proposed cylindrical TPV representation followed by 2D image backbones. 
We can then effectively restore the 3D structured representation from the processed TPV to obtain high-resolution 3D occupancy predictions.

\section{Proposed Approach} \label{method}

\subsection{Efficient Representation for Point Cloud}
Given a point cloud $\mathcal{R}^{\mathbf{P}\in N\times C_{in}}$, 3D semantic occupancy prediction aims to predict a semantic label for each voxel in the 3D space.
LiDAR segmentation can be seen as a sparse special case of 3D semantic occupancy prediction, where only the semantic class for each scanned point is required.
The best-performing methods usually employ the voxel representation\cite{voxelnet, centerpoint} to describe a 3D scene with dense cubic features $\mathbf{V} \in \mathcal{R}^{H\times W\times D\times C}$, where $H$, $W$, $D$ represent the spatial resolution of voxels and C represents the channel dimension.
However, as its computation and storage complexity are proportional to $O(HWD)$, only a low-resolution voxel representation can be learned.

2D-projection-based methods~\cite{polarnet, amvnet, rangenet++}  address this by projecting point clouds onto a 2D plane, where a popular choice is the range view~\cite{salsanext, rangenet++, rangevit}.
By transforming point clouds to the $r-\theta-\phi$ plane, they compress them along the r-axis in the spherical coordinate system to obtain $\theta-\phi$ 2D plane features.
Since the features are encoded on the 2D plane, the range-view representation only needs to use the 2D backbone without complex 3D convolution operations.
However, it loses radial information when compressing the r-axis, resulting in the inability to recover dense features of 3D scenes from range-view representation.
Hence, it is not suitable for 3D dense prediction tasks, such as 3D semantic occupancy prediction.

\begin{figure}[t] 
\centering
\includegraphics[width=0.45\textwidth]{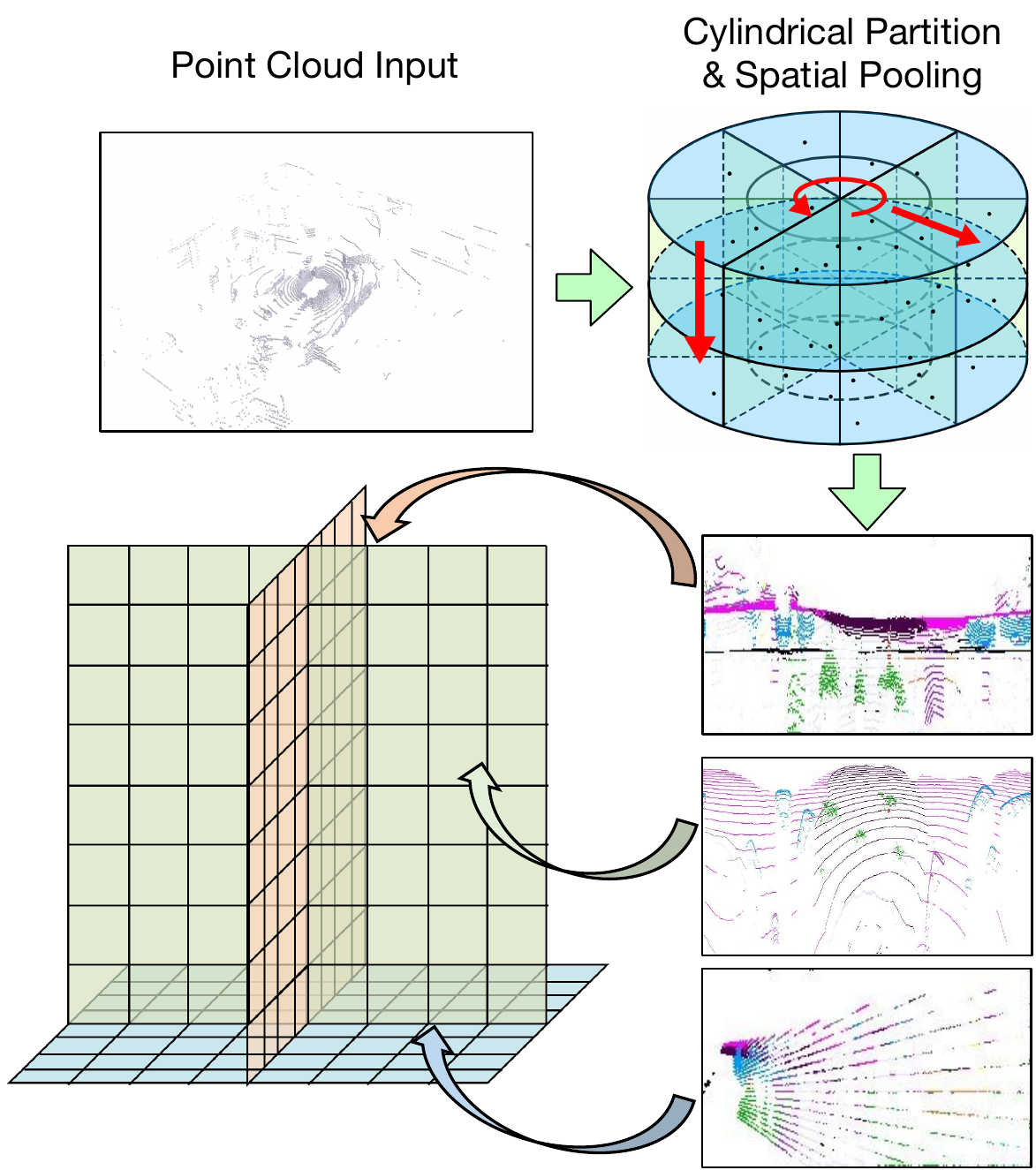}
	\vspace{-3mm}
\caption{Illustration of the proposed cylindrical tri-perspective view. 
We employ cylindrical partition and spatial pooling to obtain the cylindrical tri-perspective view to effectively represent LiDAR point clouds using 2D planes while preserving the 3D structure.
}
\label{fig:TPV}
	\vspace{-5mm}
\end{figure}

For an efficient 3D-structure-preserving 2D processing of point clouds, we propose PointOcc, which introduces the Tri-Perspective View (TPV)~\cite{tpvformer} to point cloud perception for the first time to maintain the capacity to model complex 3D scenes while avoiding cubic complexity, as shown in Figure~\ref{fig:TPV}.

\subsection{PointOcc}

\begin{figure*}
    \centering
    \includegraphics[width=\linewidth]{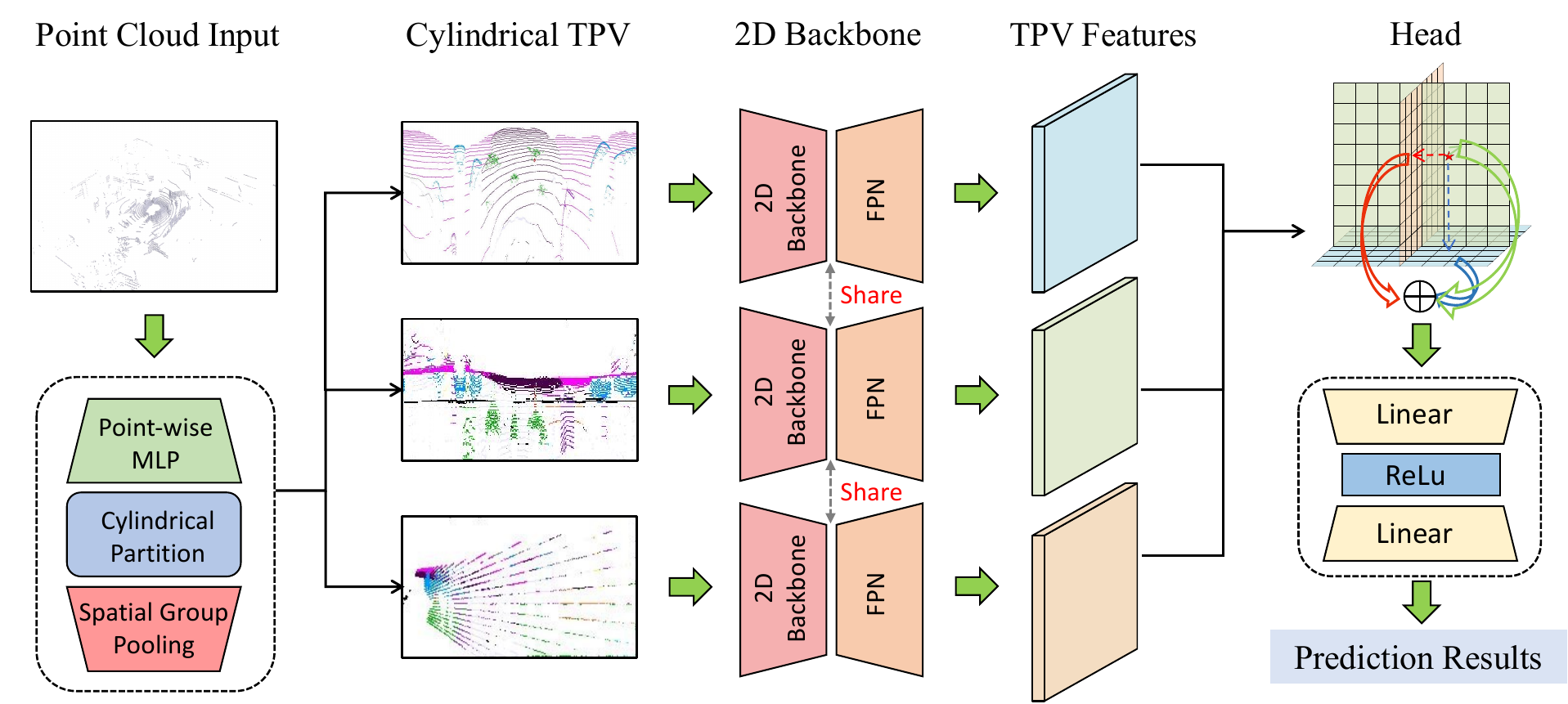}
	\vspace{-7mm}
    \caption{Illustration of the overall architecture of the proposed PointOcc.
    We first transform the LiDAR point cloud into the cylindrical TPV representation and then use 2D backbones to process it.
    }
    \label{fig:overall}
	\vspace{-7mm}
\end{figure*}

\textbf{Overall Architecture}. 
As shown in Figure~\ref{fig:overall}, the overall framework of PointOcc is composed of three parts: LiDAR projector, TPV encoder-decoder, and task-specific head. 
Given a point cloud input $\mathbf{P}$, the LiDAR projector firstly encodes point features with a point-wise MLP (\ref{encode point feature}), and then performs cylindrical projection (\ref{cylinderical partition}) and spatial pooling (\ref{spatial pooling}) to obtain cylindrical TPV inputs, i.e., three 2D perpendicular planes in the cylindrical coordinate system: 
\begin{equation}
    \mathbf{F}_{HW}, \mathbf{F}_{WD}, \mathbf{F}_{DH} = \text{LidarProjector} (\mathbf{P}), \end{equation}
where \begin{equation}
    \begin{aligned}
        \mathbf{F}_{HW} \in \mathcal{R}^{\mathcal{H}\times \mathcal{W}\times\mathcal{C}}, \space
        \mathbf{F}_{WD} \in \mathcal{R}^{\mathcal{W}\times \mathcal{D}\times\mathcal{C}}, \space
        \mathbf{F}_{DH} \in \mathcal{R}^{\mathcal{D}\times \mathcal{H}\times\mathcal{C}}.
    \end{aligned}
\end{equation}
$\mathcal{H}$,$\mathcal{W}$,$\mathcal{D}$ are the spatial resolution of TPV representation, and $\mathcal{C}$ denotes the channel dimension of the TPV inputs. 

Then TPV plane features can be encoded by the TPV encoder-decoder, which contains a 2D backbone and a Feature Pyramid Network(FPN) to extract and aggregate multi-scale features. 
Notably, three TPV planes are encoded by the same 2D backbone and FPN subsequently:
\begin{equation}
    \begin{aligned}
        \mathbf{T}_{HW} &= \text{TPVEncoder-Decoder}(\mathbf{F}_{HW}), \\ 
        \mathbf{T}_{WD} &= \text{TPVEncoder-Decoder}(\mathbf{F}_{WD}), \\ 
        \mathbf{T}_{DH} &= \text{TPVEncoder-Decoder}(\mathbf{F}_{DH}). \\ 
    \end{aligned}
\end{equation}
Finally, we can convert the TPV representation into point and voxel features in 3D space, i.e. $\mathbf{t}_{Point}$ and $\mathbf{t}_{Voxel}$,  through feature sampling and aggregation as defined in (\ref{TPV coordinate transformation}), (\ref{TPV sampling and aggregation}).
Specifically, for each point in the 3D space, we project it onto the three planes and perform linear interpolation if necessary to obtain a feature for each view.
We then add the three features and employ a light head for classification.

Semantic labels can be predicted by the task-specific head with point and voxel features:
\begin{equation}
    \begin{aligned}
        \text{{Pred.}{Point}} &= \text{TaskHead}(\mathbf{t}_{Point});\\
        \text{{Pred.}{Voxel}} &= \text{TaskHead}(\mathbf{t}_{Voxel}).\\
    \end{aligned}
\end{equation}
Note that PointOcc can perform two kinds of dense prediction tasks, i.e. LiDAR segmentation and semantic occupancy prediction. 

Since TPV features are represented by three planes with the shape of ($\mathcal{H}\mathcal{W}$,$\mathcal{W}\mathcal{D}$,$\mathcal{D}\mathcal{H}$), it greatly reduce the computation and storage complexity to $O(\mathcal{H}\mathcal{W}+\mathcal{W}\mathcal{D}+\mathcal{D}\mathcal{H})$ while preserving the capacity to model complex 3D scenes. 
Moreover, TPV features can be encoded by the 2D backbone, which brings the possibility to take advantage of the image-pretrained 2D backbone.

\textbf{Point cloud to Cylindrical TPV}.
Before encoding TPV features with the 2D backbone, the point cloud must be transformed into three-plane features through voxelization and spatial pooling. 
A conventional approach is to split the point cloud into uniform cubes and perform pooling along three axes in the Cartesian coordinate system to get the front, top, and side views, as TPVFormer\cite{tpvformer} does. 
However, as mentioned by Cylinder3D\cite{cylinder3D}, the lidar point cloud in outdoor scenes possesses the property of varying density.
Nearby areas have denser points, whereas points farther away are sparser. 
As a result, uniform cubic partition in the full space creates an irregular distribution of points in voxels, resulting in numerous empty voxels and a few voxels with significantly dense points. 
Numerous empty voxels lead to wasted computation when encoding TPV features, and voxels with dense points may lose fine-grained information during voxelization.
Thus, the regular cubic partition may impose restrictions on the model's performance.

To address this, we utilize a cylindrical partition instead. 
The cylinder coordinate system applies a larger grid size in more distant regions, which is in accordance with the density distribution of the point cloud.
This balances the distribution of points in voxels. 
Specifically, given a point cloud input $\mathbf{P}\in \mathcal{R}^{N\times C_{in}}$, its Cartesian coordinates are first extracted and transformed into cylinder coordinates $\mathbf{P}_{cor}\in \mathcal{R}^{N\times3}$, and a point-wise MLP is used to encode point cloud features $\mathbf{P}_{fea} \in \mathcal{R}^{N\times C}$:
\begin{equation} \label{encode point feature}
    \begin{aligned}
        \mathbf{P}_{cor} &= \mathcal{T}_{Cylinder}(\mathbf{P}) \\
        \mathbf{P}_{fea} &= \mathcal{M}(\mathbf{P})
    \end{aligned}
\end{equation}
where $\mathcal{T}_{Cylinder}$ denotes the transformation from the Cartesian coordinates to the cylinder coordinates, and $\mathcal{M}$ denotes the point-wise MLP.

Then voxel features can be obtained through voxelization, where MaxPooling is performed to the features of points falling within the same voxel:
\begin{equation} \label{cylinderical partition}
    \mathbf{V} = \mathcal{V}(\mathbf{P}_{fea}, \mathbf{P}_{cor}), \mathbf{V} \in \mathcal{R}^{H\times W\times D\times C}
\end{equation}
where $H$,$W$,$D$ are the spatial resolution of cylindrical partition, indicating the radius, angle and height, respectively. 
C is the channel dimension, and $\mathcal{V}$ denotes voxelization with features and coordinates of points.

In order to obtain three-plane features of the point cloud and subsequently encode TPV features, we apply spatial pooling to the voxel features along the three axes.
Given that MaxPooling along the axes in the full space may cause the loss of the fine-grained geometry details, we propose spatial group pooling to strike a balance between performance and efficiency.
Specifically, we split the voxels along the pooling axis into $K$ groups, and perform MaxPooling in each group, respectively. 
Then the features of $K$ groups are concatenated along the channel dimension and mapped back to the C channel dimension using a two-layer MLP:
\begin{equation} \label{spatial pooling}
    \begin{aligned}
        \mathbf{F}_{HW} &= \mathcal{M}_{HW}(\text{Concat}(\{\text{Pooling}_{\{D,i\}}(\mathbf{V})\}_{i=1}^K))\\
        \mathbf{F}_{WD} &= \mathcal{M}_{WD}(\text{Concat}(\{\text{Pooling}_{\{H,i\}}(\mathbf{V})\}_{i=1}^K))\\
        \mathbf{F}_{DH} &= \mathcal{M}_{DH}(\text{Concat}(\{\text{Pooling}_{\{W,i\}}(\mathbf{V})\}_{i=1}^K))\\
    \end{aligned}
\end{equation}
where $\mathbf{F}_{HW}\in \mathcal{R}^{H\times W\times C}$, $\mathbf{F}_{WD}\in \mathcal{R}^{W\times D\times C}$, $\mathbf{F}_{DH}\in \mathcal{R}^{D\times H\times C}$, $\{Pooling_{\{X,i\}}(V)\}_{i=1}^K$ denotes performing spatial group pooling to $\mathbf{V}$ along the X-axis with the group size of $K$, and $Concat$ denotes concatenation operation.
Then the three-plane features can be further encoded into TPV features by the 2D backbone.

Note that Pooling voxels along three axes under the cylindrical partition has a clear physical interpretation. 
$\mathbf{F}_{HW}$ is obtained by compressing the cylindrical space along the z-axis, so it can be regarded as a circular BEV plane.
Similarly, $\mathbf{F}_{WD}$ is obtained by compressing the cylindrical space along the r-axis, which can be considered as the range-view plane in the cylinder coordinate system. 
And $\mathbf{F}_{DH}$ can be thought of as complementary to the other two planes. 
By incorporating the cylindrical partition and TPV representation, the BEV and range-view representation are combined and extended in our method.

\textbf{TPV Encoder-Decoder}.
To encode TPV features, we adopt a 2D image backbone as the TPV encoder to process each plane $\mathbf{F}$, obtaining corresponding multi-scale features.
Then a Feature Pyramid Network (FPN)~\cite{FPN} is utilized as the TPV decoder to aggregate multi-level features and restore high-resolution TPV representation $\mathbf{T}_{HW}$, $\mathbf{T}_{WD}$, and $\mathbf{T}_{DH}$.
Note that both the TPV encoder and decoder share weights between the three planes, enabling the implicit interaction between the TPV plane features.

\textbf{TPV to 3D}.
Given a query point at $(x,y,z)$ in the real world, we first perform the coordinate transformation from the real world to the TPV view:
\begin{equation} \label{TPV coordinate transformation}
    [h, w, d] = \mathcal{T}_{TPV}([x,y,z])
\end{equation}
Then the TPV features can be sampled at the corresponding locations on the TPV planes, which are subsequently aggregated by summation to obtain the final TPV representation of the point:
\begin{equation} \label{TPV sampling and aggregation}
    \begin{aligned}
        \mathbf{t}_{hw} &= \mathcal{S}(\mathbf{T}_{HW}, [h,w]),
        \mathbf{t}_{wd} = \mathcal{S}(\mathbf{T}_{WD}, [w,d]), \\
        \mathbf{t}_{dh} &= \mathcal{S}(\mathbf{T}_{DH}, [d,h]),
        \mathbf{t}_{Point} = \mathbf{t}_{hw} + \mathbf{t}_{wd} + \mathbf{t}_{dh},
    \end{aligned}    
\end{equation}
where $\mathcal{S}$ denotes sampling features from TPV planes according to projected locations, and $\mathbf{t}_{Point}$ is the final TPV feature of the query point. 
This can be easily extended to the query voxel by simply replacing the coordinate of the point with the center of the voxel, thus generating a dense representation of the 3D scene. 

\textbf{Semantic Occupancy Prediction Head}.
For dense voxel prediction, the center of each voxel is used as the query position to obtain features from TPV planes as defined in (\ref{TPV coordinate transformation}) and (\ref{TPV sampling and aggregation}).
We adopt a simple two-layer MLP as the segmentation head.

\textbf{Lidar Segmentation Head}.
To conduct point segmentation, per-point features are first obtained from TPV planes as defined in (\ref{TPV coordinate transformation}) and (\ref{TPV sampling and aggregation}).
Then a lightweight segmentation head is applied to predict point-wise semantic labels.

\definecolor{LightGrey}{rgb}{.9,.9,.9}
\definecolor{White}{rgb}{1.,0.,1.}
\definecolor{first}{rgb}{.8,.0,.0}
\definecolor{second}{rgb}{.0,.6,.0}
\definecolor{third}{rgb}{.0,.0,.8}

\definecolor{nbarrier}{RGB}{255, 120, 50}
\definecolor{nbicycle}{RGB}{255, 192, 203}
\definecolor{nbus}{RGB}{255, 255, 0}
\definecolor{ncar}{RGB}{0, 150, 245}
\definecolor{nconstruct}{RGB}{0, 255, 255}
\definecolor{nmotor}{RGB}{200, 180, 0}
\definecolor{npedestrian}{RGB}{255, 0, 0}
\definecolor{ntraffic}{RGB}{255, 240, 150}
\definecolor{ntrailer}{RGB}{135, 60, 0}
\definecolor{ntruck}{RGB}{160, 32, 240}
\definecolor{ndriveable}{RGB}{255, 0, 255}
\definecolor{nother}{RGB}{139, 137, 137}
\definecolor{nsidewalk}{RGB}{75, 0, 75}
\definecolor{nterrain}{RGB}{150, 240, 80}
\definecolor{nmanmade}{RGB}{213, 213, 213}
\definecolor{nvegetation}{RGB}{0, 175, 0}

\definecolor{car}{rgb}{0.39215686, 0.58823529, 0.96078431}
\definecolor{bicycle}{rgb}{0.39215686, 0.90196078, 0.96078431}
\definecolor{motorcycle}{rgb}{0.11764706, 0.23529412, 0.58823529}
\definecolor{truck}{rgb}{0.31372549, 0.11764706, 0.70588235}
\definecolor{other-vehicle}{rgb}{0.39215686, 0.31372549, 0.98039216}
\definecolor{person}{rgb}{1.        , 0.11764706, 0.11764706}
\definecolor{bicyclist}{rgb}{1.        , 0.15686275, 0.78431373}
\definecolor{motorcyclist}{rgb}{0.58823529, 0.11764706, 0.35294118}
\definecolor{road}{rgb}{1.        , 0.        , 1.        }
\definecolor{parking}{rgb}{1.        , 0.58823529, 1.        }
\definecolor{sidewalk}{rgb}{0.29411765, 0.        , 0.29411765}
\definecolor{other-ground}{rgb}{0.68627451, 0.        , 0.29411765}
\definecolor{building}{rgb}{1.        , 0.78431373, 0.        }
\definecolor{fence}{rgb}{1.        , 0.47058824, 0.19607843}
\definecolor{vegetation}{rgb}{0.        , 0.68627451, 0.        }
\definecolor{trunk}{rgb}{0.52941176, 0.23529412, 0.        }
\definecolor{terrain}{rgb}{0.58823529, 0.94117647, 0.31372549}
\definecolor{pole}{rgb}{1.        , 0.94117647, 0.58823529}
\definecolor{traffic-sign}{rgb}{1.        , 0.        , 0.    }   

\makeatletter
\newcommand{\car@semkitfreq}{3.92}
\newcommand{\bicycle@semkitfreq}{0.03}
\newcommand{\motorcycle@semkitfreq}{0.03}
\newcommand{\truck@semkitfreq}{0.16}
\newcommand{\othervehicle@semkitfreq}{0.20}
\newcommand{\person@semkitfreq}{0.07}
\newcommand{\bicyclist@semkitfreq}{0.07}
\newcommand{\motorcyclist@semkitfreq}{0.05}
\newcommand{\road@semkitfreq}{15.30}  %
\newcommand{\parking@semkitfreq}{1.12}
\newcommand{\sidewalk@semkitfreq}{11.13}  %
\newcommand{\otherground@semkitfreq}{0.56}
\newcommand{\building@semkitfreq}{14.1}  %
\newcommand{\fence@semkitfreq}{3.90}
\newcommand{\vegetation@semkitfreq}{39.3}  %
\newcommand{\trunk@semkitfreq}{0.51}
\newcommand{\terrain@semkitfreq}{9.17} %
\newcommand{\pole@semkitfreq}{0.29}
\newcommand{\trafficsign@semkitfreq}{0.08}
\newcommand{\semkitfreq}[1]{{\csname #1@semkitfreq\endcsname}}

\newcommand{\barrier@nuscenesfreq}{11.79}
\newcommand{\bicycle@nuscenesfreq}{0.18}
\newcommand{\bus@nuscenesfreq}{5.83}
\newcommand{\car@nuscenesfreq}{48.27}
\newcommand{\construction@nuscenesfreq}{1.92}
\newcommand{\motorcycle@nuscenesfreq}{0.54}
\newcommand{\pedestrian@nuscenesfreq}{2.93}
\newcommand{\trafficcone@nuscenesfreq}{0.93}
\newcommand{\trailer@nuscenesfreq}{6.22}
\newcommand{\truck@nuscenesfreq}{20.07}
\newcommand{\driveable@nuscenesfreq}{28.64}
\newcommand{\other@nuscenesfreq}{0.77}
\newcommand{\sidewalk@nuscenesfreq}{6.34}
\newcommand{\terrain@nuscenesfreq}{6.35}
\newcommand{\manmade@nuscenesfreq}{16.10}
\newcommand{\vegetation@nuscenesfreq}{11.08}
\newcommand{\nuscenesfreq}[1]{{\csname #1@nuscenesfreq\endcsname}}

\section{Experiments}

We evaluate our method on the OpenOccupancy~\cite{openoccupancy} benchmark for 3D semantic occupancy prediction and the Panoptic nuScenes~\cite{nuscenes} benchmark for LiDAR segmentation.

\begin{table*}[t] \label{occupancy prediction results}
	\footnotesize
	\setlength{\tabcolsep}{0.008\linewidth}
	\caption{\textbf{3D Semantic occupancy prediction results on nuScenes validation set~\cite{nuscenes}.} The C, L, and D denotes camera, LiDAR, and depth, respectively. Our PointOcc achieves better performance than all previous methods based on all input modalities.}
	\vspace{-3mm}
	\centering
	\begin{tabular}{c | c | c c | c c c c c c c c c c c c c c c c}
		\toprule
		Method & Input & IoU & mIoU
		& \rotatebox{90}{\textcolor{nbarrier}{$\blacksquare$} barrier}
		& \rotatebox{90}{\textcolor{nbicycle}{$\blacksquare$} bicycle}
		& \rotatebox{90}{\textcolor{nbus}{$\blacksquare$} bus}
		& \rotatebox{90}{\textcolor{ncar}{$\blacksquare$} car}
		& \rotatebox{90}{\textcolor{nconstruct}{$\blacksquare$} const. veh.}
		& \rotatebox{90}{\textcolor{nmotor}{$\blacksquare$} motorcycle}
		& \rotatebox{90}{\textcolor{npedestrian}{$\blacksquare$} pedestrian}
		& \rotatebox{90}{\textcolor{ntraffic}{$\blacksquare$} traffic cone}
		& \rotatebox{90}{\textcolor{ntrailer}{$\blacksquare$} trailer}
		& \rotatebox{90}{\textcolor{ntruck}{$\blacksquare$} truck}
		& \rotatebox{90}{\textcolor{ndriveable}{$\blacksquare$} drive. suf.}
		& \rotatebox{90}{\textcolor{nother}{$\blacksquare$} other flat}
		& \rotatebox{90}{\textcolor{nsidewalk}{$\blacksquare$} sidewalk}
		& \rotatebox{90}{\textcolor{nterrain}{$\blacksquare$} terrain}
		& \rotatebox{90}{\textcolor{nmanmade}{$\blacksquare$} manmade}
		& \rotatebox{90}{\textcolor{nvegetation}{$\blacksquare$} vegetation}
		\\
            \midrule

        MonoScene~\cite{monoscene} & C & 18.4 & 6.9 & 7.1 & 3.9 & 9.3 & 7.2 & 5.6 & 3.0 & 5.9 & 4.4 & 4.9 & 4.2 & 14.9 & 6.3 & 7.9 & 7.4 & 10.0 & 7.6 \\

        TPVFormer~\cite{tpvformer} & C & 15.3 & 7.8 & 9.3 & 4.1 & 11.3 & 10.1 & 5.2 & 4.3 & 5.9 & 5.3 & 6.8 & 6.5 & 13.6 & 9.0 & 8.3 & 8.0 & 9.2 & 8.2 \\

        3DSketch~\cite{3dsketch} & \makecell{C\&D} & 25.6 & 10.7 & 12.0 & 5.1 & 10.7 & 12.4 & 6.5 & 4.0 & 5.0 & 6.3 & 8.0 & 7.2 & 21.8 & 14.8 & 13.0 & 11.8 & 12.0 & 21.2 \\

        AICNet~\cite{aicnet} & \makecell{C\&D} & 23.8 & 10.6 & 11.5 & 4.0 & 11.8 & 12.3 & 5.1 & 3.8 & 6.2 & 6.0 & 8.2 & 7.5 & 24.1 & 13.0 & 12.8 & 11.5 & 116. & 20.2 \\

        LMSCNet~\cite{lmscnet} & L & 27.3 & 11.5 & 12.4 & 4.2 & 12.8 & 12.1 & 6.2 & 4.7 & 6.2 & 6.3 & 8.8 & 7.2 & 24.2 & 12.3 & 16.6 & 14.1 & 13.9 & 22.2 \\

        JS3C-Net~\cite{js3c} & L & 30.2 & 12.5 & 14.2 & 3.4 & 13.6 & 12.0 & 7.2 & 4.3 & 7.3 & 6.8 & 9.2 & 9.1 & 27.9 & 15.3 & 14.9 & 16.2 & 14.0 & 24.9 \\
        
		C-CONet~\cite{openoccupancy} & C & 20.1 & 12.8 & 13.2 & 8.1 & 15.4 & 17.2 & 6.3 & 11.2 & 10.0 & 8.3 & 4.7 & 12.1 & 31.4 & 18.8 & 18.7 & 16.3 & 4.8 & 8.2  \\
        
		L-CONet~\cite{openoccupancy} & L & 30.9 & 15.8 & 17.5 & 5.2 & 13.3 & 18.1 & 7.8 & 5.4 & 9.6 & 5.6 & 13.2 & 13.6 & 34.9 & 21.5 & 22.4 & 21.7 & 19.2 & 23.5  \\
		
		M-CONet~\cite{openoccupancy} & \makecell{C\&L} & 29.5 & 20.1 & 23.3 & 13.3 & \textbf{21.2} & 24.3 & \textbf{15.3} & 15.9 & 18.0 & 13.3 & 15.3 & 20.7 & 33.2 & 21.0 & 22.5 & 21.5 & 19.6 & 23.2   \\

            \midrule

		PointOcc (ours) & L & \textbf{34.1} & \textbf{23.9} & \textbf{24.9} & \textbf{19.0} & 20.9 & \textbf{25.7} & 13.4 & \textbf{25.6} & \textbf{30.6} & \textbf{17.9} & \textbf{16.7} & \textbf{21.2} & \textbf{36.5} & \textbf{25.6} & \textbf{25.7} & \textbf{24.9} & \textbf{24.8} & \textbf{29.0}  \\ 
  
		\bottomrule
	\end{tabular}
	\label{tab: occ result}
	\vspace{-5mm}
\end{table*}

\subsection{Task Description}
\textbf{3D Semantic Occupancy Prediction}
The 3D semantic occupancy prediction recently has been a popular task of 3D scene perception for autonomous driving, which requires assigning semantic labels to all regions in the full space. In the case of OpenOccupancy~\cite{openoccupancy}, the perceptive range is from [-51.2m, -51.2m, -5m] to [51.2m, 51.2m, 3m], and the voxel size is 0.2m, resulting in a volume of $512\times512\times40$ voxels for occupancy prediction.
TPV features of every voxel are obtained by querying the voxel center and further utilized to predict semantic labels.
As for the evaluation metric, we follow OpenOccupany~\cite{openoccupancy} to utilize the semantic metric $\mathbf{mIoU}$ and the geometry metric $\mathbf{IoU}$. 

\textbf{Lidar Segmentation}
The lidar segmentation task requires assigning semantic labels to every input point.
As discussed in Section~\ref{method}, it can be achieved by querying every point in TPV planes to obtain the corresponding TPV feature, where the semantic label can be predicted by the segmentation head.
To evaluate the proposed method, we follow the official guidance to adopt mean Intersection-over-Union ($\mathbf{mIoU}$) as the evaluation metric.

\begin{table*}[t]
	\footnotesize
	\setlength{\tabcolsep}{0.0095\linewidth}
	\caption{\textbf{LiDAR segmentation results on nuScenes validation set~\cite{nuscenes}.} PointOcc achieves better performance than all 2D-based methods and is comparable to voxel-based methods.}
	\vspace{-2mm}
	\centering
\begin{tabular}{l c c c c c c c c c c c c c c c c c}
		\toprule
		Method & mIoU
		& \rotatebox{90}{\textcolor{nbarrier}{$\blacksquare$} barrier}
		& \rotatebox{90}{\textcolor{nbicycle}{$\blacksquare$} bicycle}
		& \rotatebox{90}{\textcolor{nbus}{$\blacksquare$} bus}
		& \rotatebox{90}{\textcolor{ncar}{$\blacksquare$} car}
		& \rotatebox{90}{\textcolor{nconstruct}{$\blacksquare$} const. veh.}
		& \rotatebox{90}{\textcolor{nmotor}{$\blacksquare$} motorcycle}
		& \rotatebox{90}{\textcolor{npedestrian}{$\blacksquare$} pedestrian}
		& \rotatebox{90}{\textcolor{ntraffic}{$\blacksquare$} traffic cone}
		& \rotatebox{90}{\textcolor{ntrailer}{$\blacksquare$} trailer}
		& \rotatebox{90}{\textcolor{ntruck}{$\blacksquare$} truck}
		& \rotatebox{90}{\textcolor{ndriveable}{$\blacksquare$} drive. suf.}
		& \rotatebox{90}{\textcolor{nother}{$\blacksquare$} other flat}
		& \rotatebox{90}{\textcolor{nsidewalk}{$\blacksquare$} sidewalk}
		& \rotatebox{90}{\textcolor{nterrain}{$\blacksquare$} terrain}
		& \rotatebox{90}{\textcolor{nmanmade}{$\blacksquare$} manmade}
		& \rotatebox{90}{\textcolor{nvegetation}{$\blacksquare$} vegetation}
		\\
        \midrule
        \textbf{2D Projection-based}\\

            RangeNet++~\cite{rangenet++} & 65.5 & 66.0 & 21.3 & 77.2 & 80.9 & 30.2 & 66.8 & 69.6 & 52.1 & 54.2 & 72.3 & 94.1 & 66.6 & 63.5 & 70.1 & 83.1 & 79.8  \\
        
		PolarNet~\cite{polarnet} & 71.0 & 74.7 & 28.2 & 85.3 & 90.9 & 35.1 & 77.5 & 71.3 & 58.8 & 57.4 & 76.1 & 96.5 & 71.1 & 74.7 & 74.0 & 87.3 & 85.7  \\

            SalsaNext~\cite{salsanext} & 72.2 & 74.8 & 34.1 & 85.9 & 88.4 & 42.2 & 72.4 & 72.2 & 63.1 & 61.3 & 76.5 & 96.0 & 70.8 & 71.2 & 71.5 & 86.7 & 84.4 \\

            RangeViT-IN21k~\cite{rangevit} & 74.8 & 75.1 & 39.0 & 90.2 & 88.4 & 48.0 & 79.2 & 77.2 & 66.4 & 65.1 & 76.7 & 96.3 & 71.1 & 73.7 & 73.9 & 88.9 & 87.1 \\

            RangeViT-CS~\cite{rangevit} & 75.2 & 75.5 & 40.7 & 88.3 & 90.1 & 49.3 & 79.3 & 77.2 & 66.3 & 65.2 & 80.0 & 96.4 & 71.4 & 73.8 & 73.8 & 89.9 & 87.2 \\

            AMVNet~\cite{amvnet} & 76.1 & \textbf{79.8} & 32.4 & 82.2 & 86.4 & \textbf{62.5} & 81.9 & 75.3 & \textbf{72.3} & \textbf{83.5} & 65.1 & \textbf{97.4} & 67.0 & \textbf{78.8} & 74.6 & \textbf{90.8} & 87.9 \\

        \midrule
        \textbf{3D Voxel-based}\\
        
    		Cylinder3D~\cite{cylinder3D} & 76.1 & 76.4 & 40.3 & 91.3 & \textbf{93.8} & 51.3 & 78.0 & 78.9 & 64.9 & 62.1 & 84.4 & 96.8 & 71.6 & 76.4 & 75.4 & 90.5 & 87.4  \\
		
            RPVNet~\cite{rpvnet} & 77.6 & 78.2 & 43.4 & \textbf{92.7} & 93.2 & 49.0 & \textbf{85.7} & 80.5 & 66.0 & 66.9 & 84.0 & 96.9 & 73.5 & 75.9 & \textbf{76.0} & 90.6 & \textbf{88.9} \\

            LidarMultiNet~\cite{lidarmultinet} & \textbf{82.0} & $-$ & $-$ & $-$ & $-$ & $-$ & $-$ & $-$ & $-$ & $-$ & $-$ & $-$ & $-$ & $-$ & $-$ & $-$ & $-$ \\

        \midrule
        \textbf{2D TPV-based} \\
        PointOcc (random) & 72.6 & 75.9 & 33.6 & 84.5 & 90.2 & 44.0 & 74.4 & 76.4 & 60.8 & 51.6 & 78.5 & 96.6 & 70.4 & 74.5 & 74.6 & 88.7 & 86.6  \\ 
		PointOcc (ImageNet-1K) & 77.9 & 78.3 & 44.5 & 92.6 & 92.2 & 56.4 & 83.6 & 80.5 & 65.2 & 69.0 & 82.4 & 97.0 & \textbf{75.0} & 76.3 & 75.1 & 90.1 & 87.8  \\ 
        PointOcc (ImageNet-21K) & 77.4 & 78.6 & \textbf{45.9} & 92.5 & 90.9 & 58.3 & 77.1 & \textbf{80.6} & 65.3 & 62.8 & \textbf{85.5} & 97.0 & 74.4 & 76.5 & 75.3 & 89.6 & 87.7  \\ 
  
		\bottomrule
	\end{tabular}
	\label{tab: seg result}
	\vspace{-6mm}
\end{table*}

\subsection{Implementation Details}
\textbf{Model Architecture}.
We adopt the same architecture in both tasks. 
For the lidar projector, we perform a cylindrical partition with the size of $(H, W, D)=(480, 360, 32)$, where three dimensions indicate the radius, angle and height, respectively.
We select $K=16$ for the group size of spatial group pooling.
In the TPVEncoder, we adopt SwinT~\cite{swin} pretrained on ImageNet-1K~\cite{deng2009imagenet} as our 2D backbone, followed by an FPN~\cite{FPN} to aggregate multi-scale features.
The TPV representation output by the TPVEncoder is with the shape of $(\mathcal{H}, \mathcal{W}, \mathcal{D}) = (240, 180, 16)$.
To further improve performance, three TPV planes are upsampled by a ratio of $s=2$ before predicting semantic labels.

\textbf{Optimization}.
During training on both two tasks, we leverage the Adam optimizer~\cite{adamw} with a weight decay of 0.01.
We adopt a cosine learning rate scheduler with a peak value of 2e-4 and a linear warm-up for the first 500 iterations~\cite{sgdr}.
In the lidar segmentation, we employ the classic cross-entropy loss~\cite{goodfellow2016deep} and lovasz-softmax loss~\cite{lovasz}.
For the occupancy prediction, we follow OpenOccupancy~\cite{openoccupancy} to additionally use an affinity loss to optimize the geometry and semantic metrics~\cite{monoscene}.
All models are trained for 24 epochs with a batch size of 8 on 8 RTX 3090 GPUs.

\textbf{Inference}.
During inference on the occupancy prediction task, voxel features with an extremely high resolution of $[512, 512, 40]$ can not be obtained directly due to memory limitations.
Instead, we first obtain voxel features with a shape of $[256, 256, 20]$ and predict the logits of all classes. 
Then we upsample the logits back to the original resolution by tri-linear interpolation as the final occupancy prediction.

\subsection{3D Semantic Occupancy Prediction Results}
We validate the effectiveness of our method on the OpenOccupancy~\cite{openoccupancy} benchmark. 
As shown in Table \ref{tab: occ result}, our PointOcc achieves better performance than all previous methods based on all input modalities.
Compared with the lidar-based L-CONet~\cite{openoccupancy}, our method improves $\mathbf{mIoU}$ and $\mathbf{IoU}$ by $\textbf{51\%}$ and $\textbf{10\%}$, respectively, which is a giant leap on performance. 
Our method even outperforms the multimodal fusion-based M-CONet~\cite{openoccupancy} by $\textbf{3.8}$ and $\textbf{4.6}$ point on $\mathbf{mIoU}$ and $\mathbf{IoU}$, respectively, demonstrating the superior ability of our method to model complex 3D scenes.

In addition, the lidar-based L-CONet uses a complex 3D convolutional network~\cite{second}, while our PointOcc uses the 2D backbone, resulting in a significant reduction in computational burden. 
Specifically, our PointOcc reduces GFLOPs by a factor of 1.37 compared to the L-CONet.
We further construct two more efficient versions of PointOcc, named PointOcc-S and PointOcc-T. 
Compared to PointOcc, the only difference is the spatial resolution of the TPV representation, which are $[240,180,16]$ and $[120,90,8]$ for PointOcc-S and PointOcc-T, respectively.
As shown in Figure \ref{fig:overview}, PointOcc-S and PointOcc-T further reduce the GFLOPs while maintaining competitive performance on occupancy prediction, which proves the effectiveness and efficiency of our method.

\subsection{Lidar Segmentation Results}
We further conduct experiments on the nuScenes~\cite{nuscenes} validation set to verify the effectiveness of our method. 
As shown in Table \ref{tab: seg result}, PointOcc achieves better performance than all 2D projection-based methods~\cite{rangenet++, polarnet, salsanext}, including existing state-of-the-art RangeVit~\cite{rangevit}. 
Note that 2D projection-based methods like RangeVit~\cite{rangevit} usually use complicated data augmentation and post-processing techniques, while our framework is simple without any complex operations, which further validates its effectiveness. 

Our method is also compared to the 3D voxel-based methods~\cite{cylinder3D, rpvnet, lidarmultinet}, which utilize voxel representation and 3D convolutional backbones. 
In the same case of cylindrical partition, our PointOcc outperforms Cylinder3D~\cite{cylinder3D} by 1.8 on $\mathbf{mIoU}$, demonstrating its superiority. 
Moreover, PointOcc surpasses 3D-based RPVNet~\cite{rpvnet} for the first time among all 2D-based methods.

\begin{table}[t]
\begin{minipage}[t]{0.485\textwidth} 
	\caption{\textbf{Different TPV planes are utilized to obtain TPV features for LiDAR segmentation.} }
	\vspace{-2mm}
		\centering
		\begin{tabular}[b]{c|c c c|c}
			\toprule
			Model & $\mathcal{T}_{HW}$ & $\mathcal{T}_{WD}$ & $\mathcal{T}_{DH}$ & mIoU 
			\\
			\midrule
			(a) & $\checkmark$ &  &  & 75.5
			\\
			(b) &  & $\checkmark$ &  & 55.5
			\\
			(c) &  &  & $\checkmark$ & 39.1
			\\
			(d) & $\checkmark$ & $\checkmark$ & $\checkmark$ & \textbf{77.9}
            \\
			\bottomrule
		\end{tabular}
	\label{tab:ablate tpv planes}
\end{minipage}
~~~~
\begin{minipage}[t]{0.485\textwidth} 
    \caption{\textbf{Freeze partial weights of ViT for LiDAR segmentation.} ATTN: attention layers. FFN: feed-forward network.}
	\vspace{-2mm}
		\centering
		\begin{tabular}[b]{c|c c|c|c}
			\toprule
                & \multicolumn{2}{c|}{Freeze} & Learnable
                \\
			Model & ATTN & FFN & Params & mIoU 
			\\
			\midrule
			(a) & $\checkmark$ &  & 24M & 76.9 
                \\
			(b) & & $\checkmark$ & 15M & 75.5
                \\
			(c) & $\checkmark$ & $\checkmark$ &  7M & 70.8
                \\
            (d) & & & 32M & \textbf{77.9}
            \\
			\bottomrule
		\end{tabular}
	\label{tab:ablate freeze model}
    \vspace{-7mm}
\end{minipage}
\end{table}

\subsection{Analysis}

\textbf{Complementary properties of three TPV planes}.
We explore the model's performance when only one of the three TPV planes is utilized to obtain TPV features. 
As shown in Table \ref{tab:ablate tpv planes}, the model utilizing only $\mathcal{T}_{HW}$ demonstrates strong performance, achieving 75.5 $\mathbf{mIoU}$ which is a little lower than the original framework.
This is reasonable since $\mathcal{T}_{HW}$ can be considered as a circular BEV plane that has the strong capability to represent the 3D scenes.
The model with $\mathcal{T}_{WD}$ or $\mathcal{T}_{DW}$ ends up with poor performance, only achieving a \textbf{mIoU} of 55.5 and 39.1 respectively, which is far from the original performance. 
However, combining the three planes into the TPV representation can achieve better performance than all the above choices, indicating that the three TPV planes are complementary to each other and can work together to represent complex 3D scenes effectively.

\textbf{Spatial resolution of TPV representation}.
We present the results of our method with different TPV resolutions. 
Note that the resolution refers to the spatial shape of the final TPV features after 2x upsampling, which is the same as the spatial resolution of the cylindrical partition. 
As shown in Table \ref{tab:ablate tpv resolution}, the performance gradually improves as the resolution grows, because TPV representation with high resolution has a better ability to model fine-grained 3D structure.

\begin{table}[t]
\begin{minipage}[t]{0.485\textwidth} 
    \caption{\textbf{Different spatial resolution of TPV representation for LiDAR segmentation.} }
	\vspace{-2mm}
		\centering
		\begin{tabular}[b]{c|c|c}
			\toprule
			Model & Spatial Resolution & mIoU 
			\\
			\midrule
			(a) & $(120, 90, 8)$ & 70.5 \\
			(b) & $(240, 180, 16)$ & 75.7 \\
			(c) & $(480, 360, 32)$ & \textbf{77.9}
            \\
			\bottomrule
		\end{tabular}
	\label{tab:ablate tpv resolution}
\end{minipage}
~~~~
\begin{minipage}[t]{0.485\textwidth} 
    \caption{\textbf{Different group size $K$ of spatial group pooling for LiDAR segmentation.} }
	\vspace{-2mm}
		\centering
		\begin{tabular}[b]{c|c|c}
			\toprule
			Model & Group Size $K$ & mIoU 
			\\
			\midrule
			(a) & 1 & 77.3 \\
			(b) & 4 & 77.5 \\
			(c) & 16 & \textbf{77.9}
            \\
			\bottomrule
		\end{tabular}
	\label{tab:ablate group size}
    \vspace{-7mm}
\end{minipage}
\end{table}

\begin{figure*}
    \centering
    \includegraphics[width=0.98\linewidth]{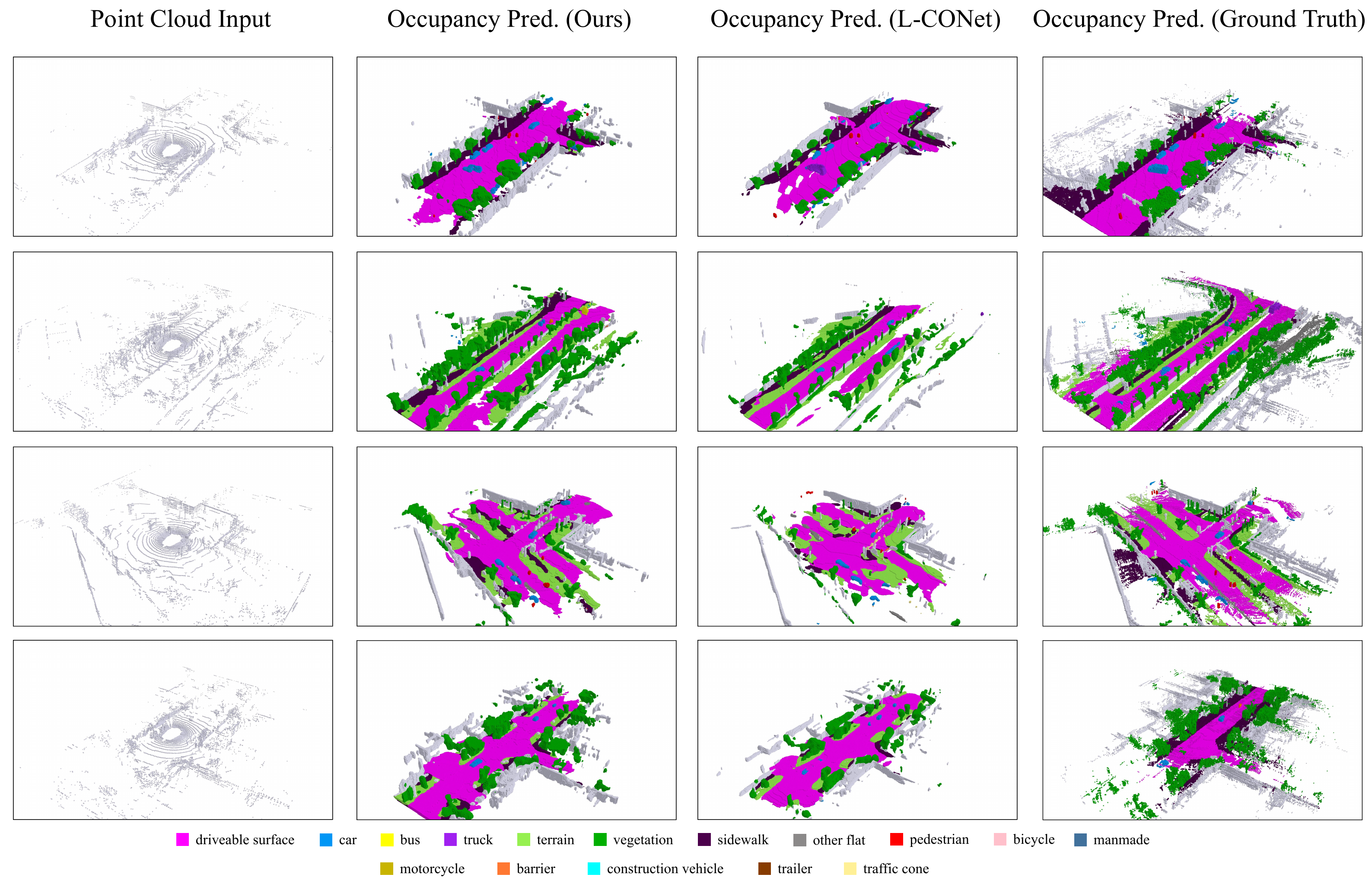}
	\vspace{-4mm}
    \caption{\textbf{Visualization results on 3D semantic occupancy prediction.} 
    Our method can generate more comprehensive and accurate prediction results than the L-CONet.
    }
    \label{fig: vis_occ}
	\vspace{-6mm}
\end{figure*}

\textbf{Spatial group pooling}.
We ablate the group size $K$ of spatial group pooling and conduct experiments on $K=1,4,16$.
As shown in Table \ref{tab:ablate group size}, $K=1$ means pooling along the axes in the full space, which may lead to a severe loss of structural information and results in degraded performance. 
When $K$ increases, the model is able to retain more fine-grained information and achieves better performance. 
Considering the computational burden and spatial resolution of TPV representation, we choose $K=16$ as the group size of spatial group pooling in the final model architecture.

\textbf{Image-pretrained 2D backbone}.
Since the 2D backbone is the core component of the whole framework, we study the performance of the model when using the 2D backbone with different initializations. 
Specifically, we conduct experiments with the vision transformer (ViT)~\cite{swin} initialized randomly and initialized with weights pretrained on ImageNet-1K and ImageNet-21K~\cite{deng2009imagenet}, respectively. 
The results are shown in Table \ref{tab: seg result}.
We find that despite the huge domain difference, using a ViT~\cite{swin} pretrained on RGB images can yield significant performance gains compared to the model initialized randomly.
And the ViT pretrained on ImageNet-1K~\cite{deng2009imagenet} performs the best.

Since ViT achieves surprisingly great performance when transferring from images to the point cloud, we further explore the impact of freezing the partial weights of ViT pre-trained on ImageNet-1K~\cite{deng2009imagenet}.
As shown in Table \ref{tab:ablate freeze model}, when freezing attention layers and FFN layers in the ViT respectively, the model can achieve performance comparable to the original structure. 
When freezing both the attention layers and FFN layers, the performance does not degrade too much, demonstrating that the ViT pre-trained on images can perform well on point-based 3D perception tasks.

\textbf{Visualizations}.
Figure~\ref{fig: vis_occ} shows the visualization comparisons on 3D semantic occupancy prediction.
We see that compared with CONet, Our PointOcc can provide a more comprehensive semantic reconstruction of the 3D scene.

\section{Conclusion}

In this paper, we have introduced the cylindrical tri-perspective view representation to the point-based model, which can model fine-grained 3D structures using an efficient 2D image backbone. 
To transform point clouds into TPV space, we have proposed cylindrical partition and spatial group pooling to maintain structural information.
Experiments on lidar segmentation and occupancy prediction show that our PointOcc can achieve better performance than all 2D-projection-based methods and is comparable to voxel-based methods.
We have demonstrated that our point-based model outperforms all other methods by a large margin on the OpenOccupancy benchmark.

\textbf{Limitations}. Despite the efficiency of the PointOcc backbone, it still needs to compute dense 3D features in the segmentation head, which limits its scalability for larger-resolution scene modeling.



\appendix

\begin{table*}[t]
	\footnotesize
	\setlength{\tabcolsep}{0.0095\linewidth}
	\caption{\textbf{LiDAR segmentation results on nuScenes test set~\cite{nuscenes}.} Despite its extremely simple structure, our PointOcc achieves better performance than all 2D-based methods and is comparable to 3D voxel-based methods}
	\vspace{-2mm}
	\centering
	\begin{tabular}{l c c c c c c c c c c c c c c c c c}
		\toprule
		Method & mIoU
		& \rotatebox{90}{\textcolor{nbarrier}{$\blacksquare$} barrier}
		& \rotatebox{90}{\textcolor{nbicycle}{$\blacksquare$} bicycle}
		& \rotatebox{90}{\textcolor{nbus}{$\blacksquare$} bus}
		& \rotatebox{90}{\textcolor{ncar}{$\blacksquare$} car}
		& \rotatebox{90}{\textcolor{nconstruct}{$\blacksquare$} const. veh.}
		& \rotatebox{90}{\textcolor{nmotor}{$\blacksquare$} motorcycle}
		& \rotatebox{90}{\textcolor{npedestrian}{$\blacksquare$} pedestrian}
		& \rotatebox{90}{\textcolor{ntraffic}{$\blacksquare$} traffic cone}
		& \rotatebox{90}{\textcolor{ntrailer}{$\blacksquare$} trailer}
		& \rotatebox{90}{\textcolor{ntruck}{$\blacksquare$} truck}
		& \rotatebox{90}{\textcolor{ndriveable}{$\blacksquare$} drive. suf.}
		& \rotatebox{90}{\textcolor{nother}{$\blacksquare$} other flat}
		& \rotatebox{90}{\textcolor{nsidewalk}{$\blacksquare$} sidewalk}
		& \rotatebox{90}{\textcolor{nterrain}{$\blacksquare$} terrain}
		& \rotatebox{90}{\textcolor{nmanmade}{$\blacksquare$} manmade}
		& \rotatebox{90}{\textcolor{nvegetation}{$\blacksquare$} vegetation}
		\\
        \midrule
        \textbf{2D Projection-based}\\

            MINet~\cite{minet} & 56.3 & 54.6 & 8.2 & 62.1 & 76.6 & 23.0 & 58.7 & 37.6 & 34.9 & 61.5 & 46.9 & 93.3 & 56.4 & 63.8 & 64.8 & 79.3 & 78.3  \\
        
    		PolarNet~\cite{polarnet} & 69.4 & 72.2 & 16.8 & 77.0 & 86.5 & 51.1 & 69.7 & 64.8 & 54.1 & 69.7 & 63.5 & 96.6 & 67.1 & 77.7 & 72.1 & 87.1 & 84.5  \\

            PolarSteam~\cite{polarstream} & 73.4 & 71.4 & 27.8 & 78.1 & 82.0 & 61.3 & 77.8 & 75.1 & 72.4 & 79.6 & 63.7 & 96.0 & 66.5 & 76.9 & 73.0 & 88.5 & 84.8  \\

            AMVNet~\cite{amvnet} & 77.3 & 80.6 & 32.0 & 81.7 & 88.9 & 67.1 & 84.3 & 76.1 & 73.5 & 84.9 & 67.3 & 97.5 & 67.4 & 79.4 & 75.5 & 91.5 & 88.7   \\

        \midrule
        \textbf{3D Voxel-based}\\

            JS3C-Net~\cite{js3c} & 73.6	& 80.1	& 26.2 & 87.8 & 84.5 & 55.2	& 72.6	& 71.3	& 66.3	& 76.8	& 71.2	& 96.8	& 64.5	& 76.9	& 74.1	& 87.5	& 86.1 \\

            SPVNAS~\cite{spvnas} & 77.4  & 80.0 & 30.0 & 91.9 & 90.8 & 64.7 & 79.0 & 75.6 & 70.9 & 81.0 & 74.6 & 97.4 & 69.2 & 80.0 & 76.1 & 89.3 & 87.1  \\

            Cylinder3D++~\cite{cylinder3D} & 77.9 & 82.8 & 33.9 & 84.3 & 89.4 & 69.6 & 79.4 & 77.3 & 73.4 & 84.6 & 69.4 & 97.7 & 70.2 & 80.3 & 75.5 & 90.4 & 87.6   \\

            AF2S3Net~\cite{af2s3net} & 78.3 & 78.9 & \textbf{52.2} & 89.9 & 84.2 & \textbf{77.4} & 74.3 & 77.3 & 72.0 & 83.9 & 73.8 & 97.1 & 66.5 & 77.5 & 74.0 & 87.7 & 86.8   \\

            DRINet++~\cite{drinet++} & 80.4 & \textbf{85.5} & 43.2 & 90.5 & \textbf{92.1} & 64.7 & 86.0 & 83.0 & 73.3 & 83.9 & \textbf{75.8} & 97.0 & \textbf{71.0} & \textbf{81.0} & \textbf{77.7} & 91.6 & \textbf{90.2}  \\

            LidarMultiNet~\cite{ye2022lidarmultinet} & \textbf{81.4} & 80.4 & 48.4 & \textbf{94.3} & 90.0 & 71.5 & \textbf{87.2} & \textbf{85.2} & \textbf{80.4} & \textbf{86.9} & 74.8 & \textbf{97.8} & 67.3 & 80.7 & 76.5 & \textbf{92.1} & 89.6   \\

        \midrule
        \textbf{2D TPV-based} \\
		PointOcc (ImageNet-1K) & 78.4 & 83.1 & 43.3 & 88.8 & 89.9 & 66.1 & 81.3 & 76.4 & 71.3 & 83.5 & 72.0 & 97.6 & 67.2 & 80.2 & 75.6 & 91.0 & 87.7  \\ 
  
		\bottomrule
	\end{tabular}
	\label{tab: seg result}
	\vspace{-6mm}
\end{table*}

\section{LiDAR Segmentation Results}

In Table~\ref{tab: seg result}, we report the performance of our PointOcc on nuScenes test set for LiDAR Segmentation.
Our method achieves better performance than all 2D Projection-based methods without any post-processing techniques.
Furthermore, our method outperforms 3D Voxel-based method AF2S3Net~\cite{af2s3net}, which is the first time for 2D Projection-based methods.
Despite the straightforward and efficient framework, our PointOcc is on par with 3D Voxel-based methods, which demonstrates the effectiveness of our method in modeling complex 3D scenes.

\section{Evalution Metric}
\textbf{Lidar Segmentation}
We follow the official guidance to adopt mean Intersection-over-Union($\mathbf{mIoU}$) as the evaluation metric:
\begin{equation} \label{miou}
    \begin{aligned}
        \mathbf{IoU}_{i} &= \frac{TP_{i}}{TP_{i} + FP_{i} + FN_{i}} \\
        \mathbf{mIoU} &= \sum_{i=1}^{cls} \mathbf{IoU}_{i}
    \end{aligned}
\end{equation}
where $TP_{i}, FP_{i}, FN_{i}$ denote true positive, false positive and false negative prediction for the i-th class, and $\mathbf{mIoU}$ is the mean of $\mathbf{IoU}$ over all classes.

\textbf{3D Semantic Occupancy Prediction}
We follow OpenOccupany~\cite{openoccupancy} to utilize the semantic metric $\mathbf{mIoU}$ defined in (\ref{miou}) and the geometry metric $\mathbf{IoU}$ defined in (\ref{iou}):
\begin{equation} \label{iou}
    \mathbf{IoU} = \frac{TP_{o}}{TP_{o} + FP_{o} + FN_{o}}
\end{equation}
where $TP_{o}, FP_{o}, FN_{o}$ represent true positive, false positive and false negative prediction for occupied voxels.

\section{LiDAR Segmentation Visualization}

Figure~\ref{fig: vis_seg} shows the visualization comparisons on the LiDAR Segmentation task.
Compared with Cylinder3D, our method can give more accurate predictions on objects like trailers and better perceive the background.

\begin{figure*}[!t]
    \centering
    \includegraphics[width=\linewidth]{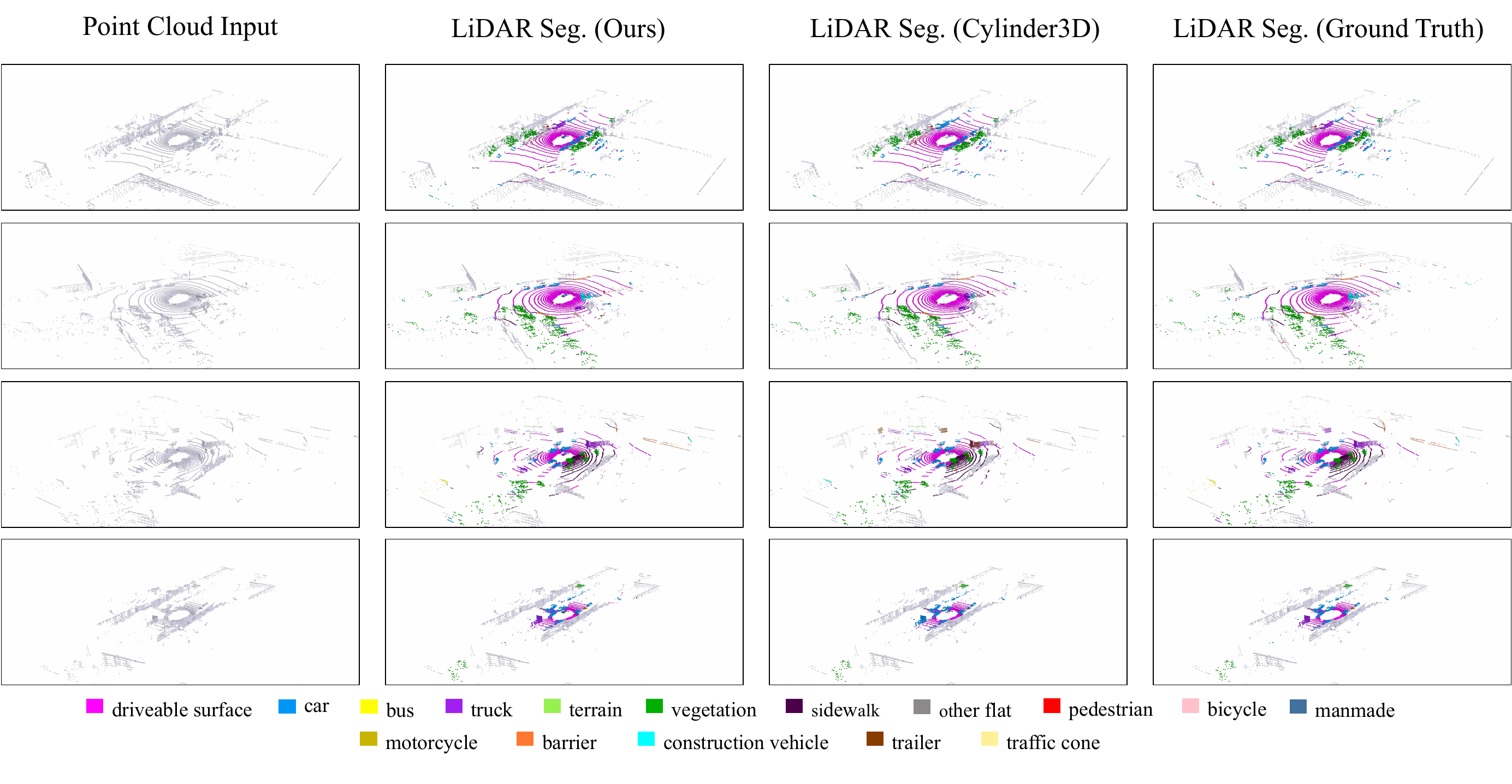}
	\vspace{-10mm}
    \caption{\textbf{Visualization results on nuScenes LiDAR segmentation.}
    }
    \label{fig: vis_seg}
\end{figure*}

\section{3D Semantic Occupancy Prediction Demo}
\begin{figure*}[!t]
    \centering
    \includegraphics[width=0.95\linewidth]{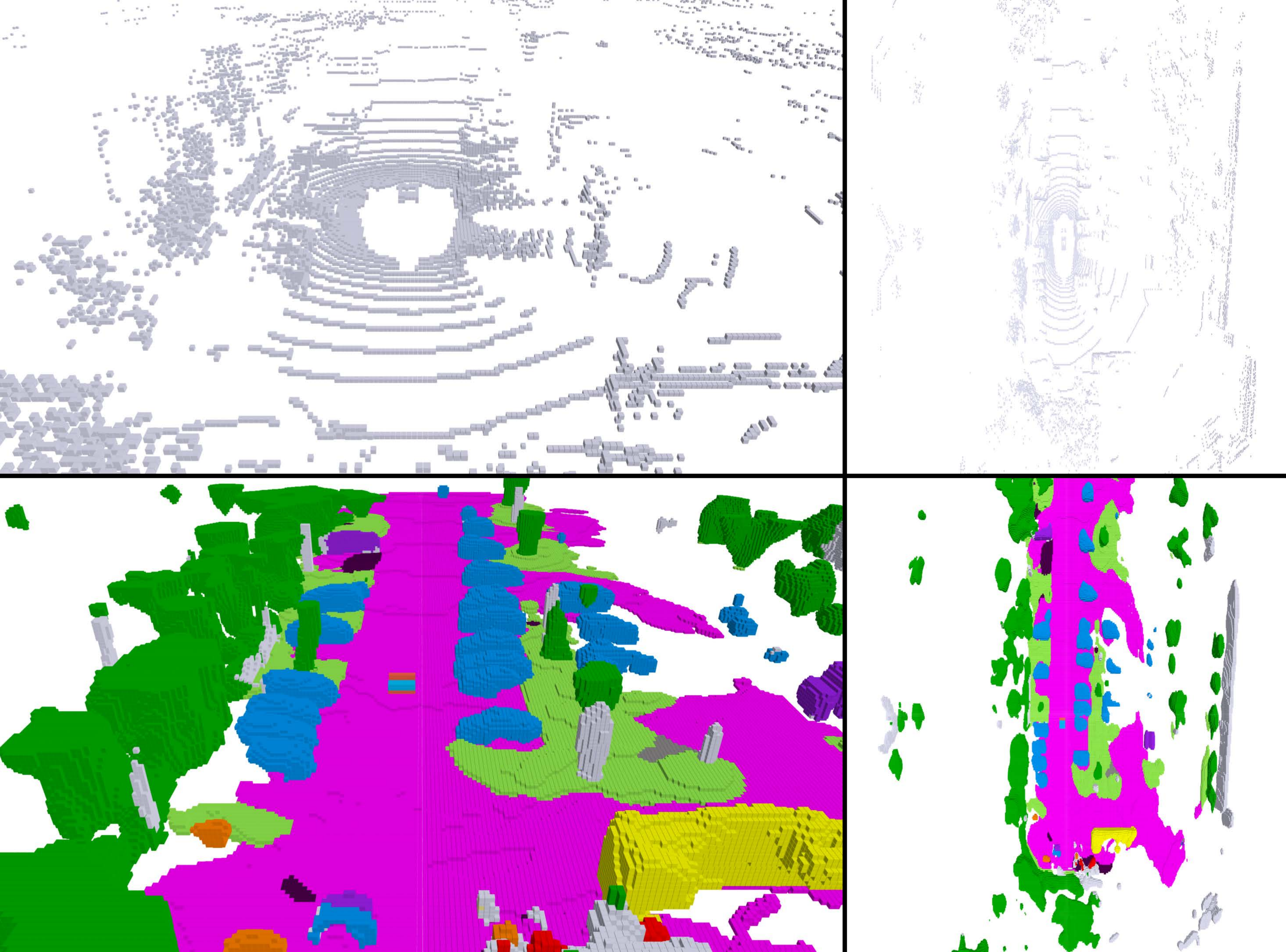}
	\vspace{-2mm}
    \caption{An image sampled from the provided demo for 3D semantic occupancy prediction on the nuScenes validation set.
    }
    \label{fig: vis_seg}
	\vspace{-2mm}
\end{figure*}

We provide a video demo\footnote{\url{https://github.com/wzzheng/PointOcc}} for 3D semantic occupancy prediction on the nuScenes validation set.
Figure~\ref{fig: vis_seg} shows an image sampled from the video demo.
For each sample, we present the LiDAR point cloud input and the occupancy prediction results from two different views.
Despite using only a single frame of point cloud input, our model achieves a comprehensive and stable perception of the 3D scene.
Especially for small objects, such as pedestrians and trees, our model is able to give accurate and robust predictions.

{\small


\begin{thebibliography}{10}\itemsep=-1pt

\bibitem{rangevit}
Angelika Ando, Spyros Gidaris, Andrei Bursuc, Gilles Puy, Alexandre Boulch, and Renaud Marlet.
\newblock Rangevit: Towards vision transformers for 3d semantic segmentation in autonomous driving.
\newblock {\em arXiv preprint arXiv:2301.10222}, 2023.

\bibitem{transfusion}
Xuyang Bai, Zeyu Hu, Xinge Zhu, Qingqiu Huang, Yilun Chen, Hongbo Fu, and Chiew-Lan Tai.
\newblock Transfusion: Robust lidar-camera fusion for 3d object detection with transformers.
\newblock In {\em CVPR}, pages 1090--1099, 2022.

\bibitem{lovasz}
Maxim Berman, Amal~Rannen Triki, and Matthew~B Blaschko.
\newblock The lov{\'a}sz-softmax loss: A tractable surrogate for the optimization of the intersection-over-union measure in neural networks.
\newblock In {\em CVPR}, pages 4413--4421, 2018.

\bibitem{nuscenes}
Holger Caesar, Varun Bankiti, Alex~H Lang, Sourabh Vora, Venice~Erin Liong, Qiang Xu, Anush Krishnan, Yu Pan, Giancarlo Baldan, and Oscar Beijbom.
\newblock nuscenes: A multimodal dataset for autonomous driving.
\newblock In {\em CVPR}, 2020.

\bibitem{monoscene}
Anh-Quan Cao and Raoul de Charette.
\newblock Monoscene: Monocular 3d semantic scene completion.
\newblock In {\em CVPR}, pages 3991--4001, 2022.

\bibitem{polarstream}
Qi Chen, Sourabh Vora, and Oscar Beijbom.
\newblock Polarstream: Streaming object detection and segmentation with polar pillars.
\newblock {\em NeurIPS}, 34:26871--26883, 2021.

\bibitem{3dsketch}
Xiaokang Chen, Kwan-Yee Lin, Chen Qian, Gang Zeng, and Hongsheng Li.
\newblock 3d sketch-aware semantic scene completion via semi-supervised structure prior.
\newblock In {\em CVPR}, pages 4193--4202, 2020.

\bibitem{voxelnext}
Yukang Chen, Jianhui Liu, Xiangyu Zhang, Xiaojuan Qi, and Jiaya Jia.
\newblock Voxelnext: Fully sparse voxelnet for 3d object detection and tracking.
\newblock {\em arXiv preprint arXiv:2303.11301}, 2023.

\bibitem{s3cnet}
Ran Cheng, Christopher Agia, Yuan Ren, Xinhai Li, and Liu Bingbing.
\newblock S3cnet: A sparse semantic scene completion network for lidar point clouds.
\newblock In {\em CoRL}, pages 2148--2161. PMLR, 2021.

\bibitem{af2s3net}
Ran Cheng, Ryan Razani, Ehsan Taghavi, Enxu Li, and Bingbing Liu.
\newblock 2-s3net: Attentive feature fusion with adaptive feature selection for sparse semantic segmentation network.
\newblock In {\em CVPR}, pages 12547--12556, 2021.

\bibitem{salsanext}
Tiago Cortinhal, George Tzelepis, and Eren Erdal~Aksoy.
\newblock Salsanext: Fast, uncertainty-aware semantic segmentation of lidar point clouds.
\newblock In {\em ISVC}, pages 207--222, 2020.

\bibitem{deng2009imagenet}
Jia Deng, Wei Dong, Richard Socher, Li-Jia Li, Kai Li, and Li Fei-Fei.
\newblock Imagenet: A large-scale hierarchical image database.
\newblock In {\em CVPR}, pages 248--255, 2009.

\bibitem{goodfellow2016deep}
Ian Goodfellow, Yoshua Bengio, and Aaron Courville.
\newblock {\em Deep learning}.
\newblock MIT press, 2016.

\bibitem{sparseconv}
Benjamin Graham, Martin Engelcke, and Laurens Van Der~Maaten.
\newblock 3d semantic segmentation with submanifold sparse convolutional networks.
\newblock In {\em CVPR}, pages 9224--9232, 2018.

\bibitem{pvkd}
Yuenan Hou, Xinge Zhu, Yuexin Ma, Chen~Change Loy, and Yikang Li.
\newblock Point-to-voxel knowledge distillation for lidar semantic segmentation.
\newblock In {\em CVPR}, pages 8479--8488, 2022.

\bibitem{tpvformer}
Yuanhui Huang, Wenzhao Zheng, Yunpeng Zhang, Jie Zhou, and Jiwen Lu.
\newblock Tri-perspective view for vision-based 3d semantic occupancy prediction.
\newblock {\em arXiv preprint arXiv:2302.07817}, 2023.

\bibitem{kprnet}
Deyvid Kochanov, Fatemeh~Karimi Nejadasl, and Olaf Booij.
\newblock Kprnet: Improving projection-based lidar semantic segmentation.
\newblock {\em arXiv preprint arXiv:2007.12668}, 2020.

\bibitem{crf}
Philipp Kr{\"a}henb{\"u}hl and Vladlen Koltun.
\newblock Efficient inference in fully connected crfs with gaussian edge potentials.
\newblock {\em NeurIPS}, 24, 2011.

\bibitem{sphericaltr}
Xin Lai, Yukang Chen, Fanbin Lu, Jianhui Liu, and Jiaya Jia.
\newblock Spherical transformer for lidar-based 3d recognition.
\newblock {\em arXiv preprint arXiv:2303.12766}, 2023.

\bibitem{pointpillars}
Alex~H Lang, Sourabh Vora, Holger Caesar, Lubing Zhou, Jiong Yang, and Oscar Beijbom.
\newblock Pointpillars: Fast encoders for object detection from point clouds.
\newblock In {\em CVPR}, 2019.

\bibitem{aicnet}
Jie Li, Kai Han, Peng Wang, Yu Liu, and Xia Yuan.
\newblock Anisotropic convolutional networks for 3d semantic scene completion.
\newblock In {\em CVPR}, pages 3351--3359, 2020.

\bibitem{minet}
Shijie Li, Xieyuanli Chen, Yun Liu, Dengxin Dai, Cyrill Stachniss, and Juergen Gall.
\newblock Multi-scale interaction for real-time lidar data segmentation on an embedded platform.
\newblock {\em RA-L}, 7(2):738--745, 2021.

\bibitem{FPN}
Tsung-Yi Lin, Piotr Dollár, Ross Girshick, Kaiming He, Bharath Hariharan, and Serge Belongie.
\newblock Feature pyramid networks for object detection.
\newblock In {\em CVPR}, 2017.

\bibitem{amvnet}
Venice~Erin Liong, Thi Ngoc~Tho Nguyen, Sergi Widjaja, Dhananjai Sharma, and Zhuang~Jie Chong.
\newblock Amvnet: Assertion-based multi-view fusion network for lidar semantic segmentation.
\newblock {\em arXiv preprint arXiv:2012.04934}, 2020.

\bibitem{swin}
Ze Liu, Yutong Lin, Yue Cao, Han Hu, Yixuan Wei, Zheng Zhang, Stephen Lin, and Baining Guo.
\newblock Swin transformer: Hierarchical vision transformer using shifted windows.
\newblock In {\em ICCV}, pages 10012--10022, 2021.

\bibitem{bevfusion}
Zhijian Liu, Haotian Tang, Alexander Amini, Xinyu Yang, Huizi Mao, Daniela Rus, and Song Han.
\newblock Bevfusion: Multi-task multi-sensor fusion with unified bird's-eye view representation.
\newblock {\em arXiv preprint arXiv:2205.13542}, 2022.

\bibitem{sgdr}
Ilya Loshchilov and Frank Hutter.
\newblock Sgdr: Stochastic gradient descent with warm restarts.
\newblock {\em arXiv preprint arXiv:1608.03983}, 2016.

\bibitem{adamw}
Ilya Loshchilov and Frank Hutter.
\newblock Decoupled weight decay regularization.
\newblock In {\em ICLR}, 2019.

\bibitem{voxeltr}
Jiageng Mao, Yujing Xue, Minzhe Niu, Haoyue Bai, Jiashi Feng, Xiaodan Liang, Hang Xu, and Chunjing Xu.
\newblock Voxel transformer for 3d object detection.
\newblock In {\em ICCV}, pages 3164--3173, 2021.

\bibitem{rangenet++}
Andres Milioto, Ignacio Vizzo, Jens Behley, and Cyrill Stachniss.
\newblock Rangenet++: Fast and accurate lidar semantic segmentation.
\newblock In {\em IROS}, pages 4213--4220, 2019.

\bibitem{gfnet}
Haibo Qiu, Baosheng Yu, and Dacheng Tao.
\newblock Gfnet: Geometric flow network for 3d point cloud semantic segmentation.
\newblock {\em arXiv preprint arXiv:2207.02605}, 2022.

\bibitem{lmscnet}
Luis Roldao, Raoul de Charette, and Anne Verroust-Blondet.
\newblock Lmscnet: Lightweight multiscale 3d semantic completion.
\newblock In {\em 3DV}, pages 111--119. IEEE, 2020.

\bibitem{pv-rcnn}
Shaoshuai Shi, Chaoxu Guo, Li Jiang, Zhe Wang, Jianping Shi, Xiaogang Wang, and Hongsheng Li.
\newblock Pv-rcnn: Point-voxel feature set abstraction for 3d object detection.
\newblock In {\em CVPR}, 2020.

\bibitem{spvnas}
Haotian Tang, Zhijian Liu, Shengyu Zhao, Yujun Lin, Ji Lin, Hanrui Wang, and Song Han.
\newblock Searching efficient 3d architectures with sparse point-voxel convolution.
\newblock In {\em ECCV}, pages 685--702, 2020.

\bibitem{kpconv}
Hugues Thomas, Charles~R Qi, Jean-Emmanuel Deschaud, Beatriz Marcotegui, Fran{\c{c}}ois Goulette, and Leonidas~J Guibas.
\newblock Kpconv: Flexible and deformable convolution for point clouds.
\newblock In {\em ICCV}, pages 6411--6420, 2019.

\bibitem{occ3d}
Xiaoyu Tian, Tao Jiang, Longfei Yun, Yue Wang, Yilun Wang, and Hang Zhao.
\newblock Occ3d: A large-scale 3d occupancy prediction benchmark for autonomous driving.
\newblock {\em arXiv preprint arXiv:2304.14365}, 2023.

\bibitem{openoccupancy}
Xiaofeng Wang, Zheng Zhu, Wenbo Xu, Yunpeng Zhang, Yi Wei, Xu Chi, Yun Ye, Dalong Du, Jiwen Lu, and Xingang Wang.
\newblock Openoccupancy: A large scale benchmark for surrounding semantic occupancy perception.
\newblock {\em arXiv preprint arXiv:2303.03991}, 2023.

\bibitem{wei2023surroundocc}
Yi Wei, Linqing Zhao, Wenzhao Zheng, Zheng Zhu, Jie Zhou, and Jiwen Lu.
\newblock Surroundocc: Multi-camera 3d occupancy prediction for autonomous driving.
\newblock In {\em ICCV}, 2023.

\bibitem{scpnet}
Zhaoyang Xia, Youquan Liu, Xin Li, Xinge Zhu, Yuexin Ma, Yikang Li, Yuenan Hou, and Yu Qiao.
\newblock Scpnet: Semantic scene completion on point cloud.
\newblock {\em arXiv preprint arXiv:2303.06884}, 2023.

\bibitem{rpvnet}
Jianyun Xu, Ruixiang Zhang, Jian Dou, Yushi Zhu, Jie Sun, and Shiliang Pu.
\newblock Rpvnet: A deep and efficient range-point-voxel fusion network for lidar point cloud segmentation.
\newblock In {\em ICCV}, pages 16024--16033, 2021.

\bibitem{js3c}
Xu Yan, Jiantao Gao, Jie Li, Ruimao Zhang, Zhen Li, Rui Huang, and Shuguang Cui.
\newblock Sparse single sweep lidar point cloud segmentation via learning contextual shape priors from scene completion.
\newblock In {\em AAAI}, volume~35, pages 3101--3109, 2021.

\bibitem{second}
Yan Yan, Yuxing Mao, and Bo Li.
\newblock Second: Sparsely embedded convolutional detection.
\newblock {\em Sensors}, 2018.

\bibitem{lidarmultinet}
Dongqiangzi Ye, Zixiang Zhou, Weijia Chen, Yufei Xie, Yu Wang, Panqu Wang, and Hassan Foroosh.
\newblock Lidarmultinet: Towards a unified multi-task network for lidar perception.
\newblock {\em arXiv preprint arXiv:2209.09385}, 2022.

\bibitem{ye2022lidarmultinet}
Dongqiangzi Ye, Zixiang Zhou, Weijia Chen, Yufei Xie, Yu Wang, Panqu Wang, and Hassan Foroosh.
\newblock Lidarmultinet: Towards a unified multi-task network for lidar perception.
\newblock {\em arXiv preprint arXiv:2209.09385}, 2022.

\bibitem{drinet++}
Maosheng Ye, Rui Wan, Shuangjie Xu, Tongyi Cao, and Qifeng Chen.
\newblock Drinet++: Efficient voxel-as-point point cloud segmentation.
\newblock {\em arXiv preprint arXiv: 2111.08318}, 2021.

\bibitem{centerpoint}
Tianwei Yin, Xingyi Zhou, and Philipp Krahenbuhl.
\newblock Center-based 3d object detection and tracking.
\newblock In {\em CVPR}, pages 11784--11793, 2021.

\bibitem{bytetrack}
Yifu Zhang, Peize Sun, Yi Jiang, Dongdong Yu, Fucheng Weng, Zehuan Yuan, Ping Luo, Wenyu Liu, and Xinggang Wang.
\newblock Bytetrack: Multi-object tracking by associating every detection box.
\newblock In {\em ECCV}, pages 1--21. Springer, 2022.

\bibitem{polarnet}
Yang Zhang, Zixiang Zhou, Philip David, Xiangyu Yue, Zerong Xi, Boqing Gong, and Hassan Foroosh.
\newblock Polarnet: An improved grid representation for online lidar point clouds semantic segmentation.
\newblock In {\em CVPR}, pages 9601--9610, 2020.

\bibitem{beverse}
Yunpeng Zhang, Zheng Zhu, Wenzhao Zheng, Junjie Huang, Guan Huang, Jie Zhou, and Jiwen Lu.
\newblock Beverse: Unified perception and prediction in birds-eye-view for vision-centric autonomous driving.
\newblock {\em arXiv preprint arXiv:2205.09743}, 2022.

\bibitem{voxelnet}
Yin Zhou and Oncel Tuzel.
\newblock Voxelnet: End-to-end learning for point cloud based 3d object detection.
\newblock In {\em CVPR}, pages 4490--4499, 2018.

\bibitem{cylinder3D}
Xinge Zhu, Hui Zhou, Tai Wang, Fangzhou Hong, Yuexin Ma, Wei Li, Hongsheng Li, and Dahua Lin.
\newblock Cylindrical and asymmetrical 3d convolution networks for lidar segmentation.
\newblock In {\em CVPR}, pages 9939--9948, 2021.

\bibitem{udnet}
Hao Zou, Xuemeng Yang, Tianxin Huang, Chujuan Zhang, Yong Liu, Wanlong Li, Feng Wen, and Hongbo Zhang.
\newblock Up-to-down network: Fusing multi-scale context for 3d semantic scene completion.
\newblock In {\em IROS}, pages 16--23. IEEE, 2021.

\end{thebibliography}
}

\end{document}